\RequirePackage{fix-cm}
\documentclass{svjour3}            
\smartqed  
\usepackage[]{graphicx}
\usepackage{subcaption}
\graphicspath{{./figs/}}
\usepackage{amsmath,amssymb}
\usepackage{hyperref}
\usepackage{multirow}
\usepackage{hhline}
\usepackage[misc,geometry]{ifsym}

\usepackage{algorithm}
\usepackage{algorithmicx}
\usepackage{algpseudocode}
\usepackage{linegoal}
\usepackage{booktabs}

\begin{document}

\title{Memory Enriched Big Bang Big Crunch Optimization Algorithm for Data Clustering
}


\author{Kayvan Bijari         \and
        Hadi Zare \and 
        Hadi Veisi\and 
        Hossein Bobarshad
}


\institute{
           H. Zare (\Letter) \at
              University of Tehran \\
              \email{h.zare@ut.ac.ir}
}

\date{Received: date / Accepted: date}

\maketitle

\begin{abstract}
Cluster analysis plays an important role in decision making  process for many knowledge-based systems. There exist a wide variety of different approaches for clustering applications including the heuristic techniques, probabilistic models, and traditional hierarchical algorithms. In this paper, a novel heuristic approach based on big bang-big crunch algorithm is proposed for clustering problems. The proposed method not only takes advantage of heuristic nature to alleviate typical clustering algorithms such as k-means, but it also benefits from the memory based scheme as compared to its similar heuristic techniques. Furthermore, the performance of the proposed algorithm is investigated based on several benchmark test functions as well as on the well-known datasets. The experimental results shows the significant superiority of the proposed method over the similar algorithms.

\end{abstract}

\section{Introduction}
\label{intro}
Unsupervised learning can be considered as an important category of machine learning techniques to uncover the interesting hidden patterns from the dataset. Unsupervised learning methods can be generally divided into clustering, dimensionality reduction, image segmentation, object recognition, and text mining techniques \cite{murphy_machine_2012}. One of the most common unsupervised learning methods is clustering. Clustering is the task of assigning a set of objects, usually vectors in a multidimensional space, into clusters in such a way that the objects in the same cluster are more similar to each other than to those in other clusters \cite{jain2010data}. Cluster analysis has attracted attention of many researchers in different fields, including data mining \cite{gan2007data}, sequence mining \cite{ahmad2010cluster}, image processing \cite{moftah2014adaptive}, feature  selection techniques \cite{moradi_hybrid_2016},  spatial data analysis \cite{gan2007data}, bioinformatics \cite{lam2013pso,dinu2014clustering}, marketing \cite{gan2007data}, city planning \cite{gan2007data}, and earthquake studies \cite{gan2007data}.

Due to not  well-defined nature of clustering problems and growing importance of clustering in a variety of different fields, lots of clustering approaches have emerged. Since clustering problem is NP-hard by its nature \cite{welch1982algorithmic}, meta-heuristic methods are proper tools to deal with this issue. Among many studies in meta-heuristic approaches for clustering, one can mention clustering algorithm based on tabu search \cite{al1995tabu}, genetic algorithms \cite{krishna1999genetic}, ant colony \cite{shelokar2004ant}, particle swarm optimization \cite{cura2012particle}, and bee colony \cite{zhang2010artificial}. Meta-heuristic methods also make use of a computationally efficient clustering algorithm and seek to find better results. One common approach which can easily be used along with meta-heuristic techniques is k-means algorithm \cite{kao2008hybridized}. The k-means is one of the well-known clustering methods. In spite of its simplicity and efficiency, it suffers from serious problems such as sensitivity to the initial position of cluster centers, empty clusters occurrence, and trapping in local optima. As a result, k-means can be prevented from finding global optimum.\par

There exists many nature inspired algorithms for solving complex optimization problems which are used extensively in research works and technological settings including a variety of  particle swarm optimization, \emph{PSO},  based approaches \cite{jordehi2015enhanced,jordehi2015enhancedPSO,zhang2015new}, ant colony based method for unsupervised learning \cite{tabakhi2014unsupervised}, algorithms which have emerged from human interactions \cite{jordehi2015brainstorm,jordehi2015seeker},  water cycle chaotic behavior  \cite{heidari2015efficient}, and  human body systems \cite{jordehi2015chaotic}.\par

One of the well-known models in theoretical physics is the Big Bang theory for illustration of the universe existence and its evolution from the past known historical spans over its large-scale evolution. A novel optimization algorithm named Big Bang-Big Crunch algorithm(BB-BC) based on these theories is first initiated in \cite{erol2006new} which have been applied in many works including economic power systems \cite{verma2013tlbo,kucuktezcan2015preventive} and signal processing \cite{tang2010big}.  On the one hand, the BB-BC algorithm has been started from theoretical concepts of cosmological physics. On the other hand, the BB-BC algorithm outperforms  a wide category of evolutionary algorithms which are very sensitive to initial solutions.  Due to its modification of the initial solution in the process of the algorithm, BB-BC is aimed at achieving the optimal solution. Thus, BB-BC could be selected as a proper choice for a variety of different optimization and intractable problems.\par
 
While the BB-BC are used in several works,  it  suffers from disadvantages such as slow convergence speed and trapping in local optimum solutions available in most of the optimization problems \cite{jordehi2014chaotic}.  The problem of converging to local optimum solutions occurred for the BB-BC approach due to greedily looking around the best ever found solutions. Due to its explorative nature, BB-BC lacks a splendid exploitation factor. Such optimization strategies should have a mechanism to make a trade-off between exploration and exploitation.

In this paper a new heuristic clustering algorithm is developed. We designed memory enriched BB-BC(ME-BB-BC) algorithm to solve the aforementioned drawbacks of the traditional BB-BC method. The proposed algorithm takes advantages of typical BB-BC algorithm and enhances it with the proper balance between exploration and exploitation factors. Proposed approach not only is capable of performing clustering task but it could also be used for other general optimization problems. Results on benchmark evaluation functions and real benchmark datasets indicate that ME-BB-BC outperforms significantly the typical BB-BC method and other meta-heuristic algorithms.

The remainder of the paper is structured as, Section \ref{sec:clustering} presents the cluster analysis. The related BB-BC algorithm is given in Section \ref{sec:BB-BC}. The proposed approach based on BB-BC algorithm is introduced in Section \ref{sec:ME-BB-BC} and is evaluated and analyzed on  benchmark functions in Section \ref{sec:evalproposedmeth}. Section \ref{sec:clustering_MBBBBC} is devoted to fundamentals of clustering using the proposed algorithm. Experimental results of the proposed clustering approach are illustrated  in Section \ref{sec:exp_res}. Finally Section \ref{sec:conc} concludes the paper with suggestions for future works.

\section{Cluster analysis}
\label{sec:clustering}
Clustering is the procedure of dividing a set of objects each explained by a vector of attributes into a finite number of clusters in a way that based on similarity functions, objects in the same cluster will be similar to each other and different from the objects in other clusters. The k-means is a computationally efficient clustering method which has been widely used \cite{gan2007data}. While it is proved that the process of the algorithm will always converge \cite{xu2008clustering}, k-means does not guarantee to find an optimal solution. The algorithm is also essentially sensitive to the primary cluster centers. Moreover, k-means also suffers from the occurrence of empty clusters during its iterations. If there is a cluster with no instance, k-means is unable to update that cluster centroids.\par

The procedure of k-means is as follows. First, k-means assigns initial values to centroids and then it continues for several iterations. In each iteration, k-means create clusters by assigning all data points to their nearest centroids and then substitute the mean of each cluster for its centroid. As stopping criterion, the number of iteration can be determined a priori or iterations continue until there is no changes in centroids. Also, a combination of these two criteria is possible.

The essential aim of a clustering algorithm, such as k-means, is to discover a proper assignment of data points to clusters and find an arrangement of $\mu_k$ vectors in a manner that sum of the squares of the distances of each data point to its nearest centroid is least. For every data point $x_n$, a corresponding set of binary indicators such as $r_{n,k} \in \{0,1\}$ is presented. Where $k \in \{1,\dots, K\}$ describes which of the $K$ clusters, the data point $x_n$ falls into. So if data point $x_n$ is assigned to cluster $k$ then $r_{n,k} = 1$, and $r_{n,j} = 0$ for $j\in\{1,\dots,K|j\neq k\}$. Therefore, an objective function can be characterized as equation \ref{eqn:costfcn}, which represents the sum of the squares of the distances of each data point to its assigned cluster(vector $\mu_k$). So the final goal is to find values for the $r_{n,k}$ and $\mu_k$ in such a way that $F$ is minimized \cite{bishop_pattern_2007}.

\begin{equation}
\label{eqn:costfcn}
F = \sum_{n=1}^{N}{\sum_{k=1}^{K}r_{n,k}\times(||X_n - \mu_k||^2)} 
\end{equation}

\section{An overview of Big Bang-Big Crunch optimization algorithm}
\label{sec:BB-BC}
There exist many theories about how the universe evolved at the first place, and the two famous theories in this regard are namely Big bang and Big crunch, \emph{BB-BC} theories. Erol and Eksin \cite{erol2006new} made use of these theories and introduced the BB-BC optimization algorithm. According to this theory, due to dissipation, Big Bang phase creates randomness along with disorder, while in the Big Crunch phase the randomly created particles will be drawn into an order. Big Bang-Big Crunch algorithm (BB-BC) starts with the big bang phase through the generation of random points around an initially chosen point and it tries to shrink the created points into a single optimized one through the center of mass in the big crunch phase. Finally, after repeating the two phases for a limited number of times, the algorithm converges to an ideal solution.

Similar to other evolutionary algorithms \cite{innovative_2013}, this method has a candidate solution where some new particles are randomly distributed around it based on a uniform manner throughout the search space. The random nature of the Big Bang is associated with the energy dissipation or transmission from an ordered state to a disordered state i.e. transmission from a candidate solution to a set of new particles(solution candidates).

The Big Bang phase is pursued by the Big Crunch phase. In this phase the new random distributed particles are drawn into an order via the center of mass. After a sequentially repetitions of Big Bang and Big Crunch steps, the distribution of randomness during Big Bang phase becomes more and more smaller and finally the algorithm converges to a solution. The process of calculating the center of mass is according to equation \eqref{eqn:centerofmass}.

\begin{equation}
\label{eqn:centerofmass}
x^c_j = \frac{\sum_{j=1}^N\frac{x_j^i}{f^i}}{\sum_{j=1}^N\frac{1}{f^i}}, \text{ for i = 1, 2, \dots, N}
\end{equation}

where $x^c_j$ is the j-th component of the center of mass, $x_j^i$ is the j-th component of i-th candidate, $f^i$ is fitness value of the i-th candidate, and finally $N$ is the number of all candidates. It should be noted that in the optimization problems, fitness($f$) of each candidate solution is calculated based on general fitness function of the optimization problem, (specially for clustering applications the equation \eqref{eqn:costfcn1} is used  in this paper). The algorithm then generates new population of particles according to equation \eqref{eqn:bigbang}.

\begin{equation}
\label{eqn:bigbang}
x^{i,new}_j = x_j^c + r  \times \frac{(x_j^{max} - x_j^{min})}{1+k}
\end{equation}

where $x_j^{i, new}$ is the new value of j-th component  of the i-th particle $x$, $r$ is a random number with a standard normal distribution, and $k$ is the iteration index. Also $x_j^{max}$ and $x_j^{min}$ are maximum and minimum acceptable values for $x_j$. Algorithm \ref{alg:bbbc} shows the pseudo code of the Big Bang-Big Crunch Algorithm.

\begin{algorithm}[!h]
	\caption{Big Bang Big Crunch Algorithm}
	\begin{algorithmic}[1]
	 \renewcommand{\algorithmicrequire}{\textbf{Input:}}
 	 \renewcommand{\algorithmicensure}{\textbf{Output:}}
 	 \Require fitness function, number of stars
 	 \Ensure  output of optimization problem
 	
 	 \\ \textit{Initialisation}:
		\State $starting\_point$ = Generate a random starting point with respect to range constraints.
		\State $num\_of\_stars$ = number of stars
		\State $dim$ = dimension of solution
		
		\Repeat
		
		\\ \textit{Big Bang Phase}:\Comment{create mass around starting point}
		\For{$i = 1$ to $num\_of\_stars$} 
		\For{$j = 1$ to $dim$}
		\State $mass[i,j] = $ generate a star based on  \eqref{eqn:bigbang}
		\EndFor
		\EndFor

		
		\\ \textit{Big Crunch Phase}:
		\State $c.o.m$ = calculate center of mass based on  \eqref{eqn:centerofmass}

		\State $starting\_point = c.o.m$ \Comment{update}
		
		\Until max number of iterations or convergence
	\end{algorithmic}
	\label{alg:bbbc}
\end{algorithm}

\section{The proposed algorithm}
\label{sec:ME-BB-BC}

In this section first we introduce the proposed algorithm and describe the important elements of the algorithm in Subsection \ref{subsec:desc-prop-method}. Then sensitivity analysis of the proposed method is investigated through benchmark functions in Subsection \ref{subsec:analysis-of-proposed}. 

\subsection{The description of proposed method}
\label{subsec:desc-prop-method}
Exploration and exploitation are two important components of evolutionary algorithms. In order to act successfully, each search algorithm needs to provide a good trade-off between these two factors. Exploration is the process of searching new solution regions of the search space, exploitation on the other hand is to search in the neighborhood of previously found solutions. As an example of exploration in the BB-BC algorithm equation \eqref{eqn:bigbang} seeks to search in the new solution regions by randomly dispatching points in solution space. It can be observed from the cycles of the BB-BC algorithm, that it greedily drops the current center of mass in favor of a better one at the end of each big bang and big crunch cycle. \par

Although the BB-BC algorithm explores the solution space greatly, it suffers from lack of proper and effective exploitation. Because of the total exploration of the search space to compute the center of masses in each iteration, the efficiency of the algorithm is sensitive to these points in each step. Moreover, it is more likely to have some local solutions in the previously computed center of masses through the process of the algorithm. If we can use these points in an efficient manner, they could yield us to more robust and better results versus the original BB-BC algorithm. \par

To make use of earlier found centers of masses and enhancing the exploitation of the  algorithm, a memory with limited size is added to the process of the algorithm in an smart vein to propose a new approach entitled as \emph{Memory Enriched Big Bang Big Crunch}, \emph{ME-BB-BC}.  We describe the  stages of the \emph{ME-BB-BC} and its procedure in the following paragraph. \par

At the end of each big bang and big crunch cycles, the calculated center of mass will be stored in the memory. Initially it is assumed that all of the saved centers of masses in the memory are good points for generating the particles forming the new center of masses. Furthermore, if the memory gets full during the algorithm, the worst solution will be substituted by the new center of mass based on the fitness of the currently saved solutions.\par

\begin{algorithm}[!h]
	\caption{Memory Enriched BB-BC}
	\begin{algorithmic}[1]
		\renewcommand{\algorithmicrequire}{\textbf{Input:}}
		\renewcommand{\algorithmicensure}{\textbf{Output:}}
		\Require fitness function, memory size, number of stars
		\Ensure  output of optimization problem
		
		\\ \textit{Initialization}:
		\State $starting\_point$ = Generate a random starting point with respect to range constraints.
		\State $solution\_memory$ = memory with size $memory\_size$
		\State $num\_of\_stars$ = number of stars
		\State $dim$ = dimension of solution
		\State $\alpha= 0.1$ \Comment{memory selection rate}
		
		\Repeat
		
		\\ \textit{Big Bang Phase}:\Comment{create mass around starting point}
		\For{$i = 1$ to $num\_of\_stars$} 
		\For{$j = 1$ to $dim$}
		\If{$rand(0,1) <= \alpha$ }
		\State $idx = rand([1,..., $memory\_size$])$ 
		\State $mass[i,j] = solution\_memory[idx, j]$ \Comment{select from memory}
		\Else
		\State $mass[i,j] = $ generate a star based on \eqref{eqn:bigbang}
		\EndIf
		\EndFor
		\EndFor
		
		
		\\ \textit{Big Crunch Phase}:
		\State $c.o.m$ = calculate center of mass based on \eqref{eqn:centerofmass}
		\If{$solution\_memory$ is not full}
		\State add $c.o.m$ into $solution\_memory$
		\Else 
		\State $worse$ = find worse solution in $solution\_memory$
		\If{$fitness(c.o.m) > fitness(worse)$}
		\State remove $worse$ from the $solution\_memory$
		\State add $c.o.m$ into $solution\_memory$
		\EndIf
		\EndIf
		
		\State $starting\_point = c.o.m$ \Comment{update}
		\State $\alpha = \alpha + 0.01\,\times\,\alpha$ \Comment{update}
		
		\Until max number of iterations or convergence
	\end{algorithmic}
	\label{alg:improvedbbbc}
\end{algorithm}

We enhance the particle generation based on a probabilistic random walk manner in such a way that the adjustable parameter  $\alpha$ is considered as the selection probability of the solutions in the memory and $1-\alpha$ is the probability for the total search space. Hence, the good aspects of the dimensions of the points in the memory are utilized in the proposed method. Moreover, the weight probabilities are linearly increased as algorithm goes by to consider more importance on the memory points. This exploitation refinement idea is similar to the decreasing values of pitch adjustment rate in harmony search algorithm \cite{mahdavi_improved_2007} and inertia weight in PSO algorithm \cite{shi_empirical_1999}. Such strategy results in better performance of the meta-heuristic algorithms, due to the fact of more exploration at beginning and more exploitative at the end in the search space of the algorithm \cite{innovative_2013}.\par
 
The details of the proposed algorithm is presented in Algorithm \ref{alg:improvedbbbc}. It describes the modified big bang phase and big crunch phases of our method. The algorithm initially depends on the max number of iterations, $\alpha$, memory size, and number of stars, lines 1-6 initialize these parameters. Lines 8-18 are devoted to the modified Big Bang phase of the proposed algorithm in which random particles around starting point are generated based on \eqref{eqn:bigbang} or selected from the solution memory, in lines 19-20 center of mass is calculated according to \eqref{eqn:centerofmass}. Furthermore, in lines 21-29 the proposed algorithm checks the solution memory and saves the new center of mass to memory. At the end of the algorithm, lines 30-31 update the parameters of the proposed method. 

Based on the pseudo code of the ME-BB-BC, the computational complexity of the algorithm is $O(r \times (n + s))$, where $r$ is the total iterations of the algorithm, $n$ is the number of stars, and $s$ is the complexity of finding the worst solution in the memory. Since the $dim$ (the dimension of optimization problem) is highly less than the $r$ and $s$  parameters,  it is skipped in the complexity analysis of the algorithm. While the computational complexity of the  proposed approach is slightly more than the traditional BB-BC method, $O(r\times n)$, the proposed method yields to significant improvement versus the earlier one. 

\subsection{Analysis of the proposed algorithm}
\label{subsec:analysis-of-proposed}
We initially investigate the sensitivity of the ME-BB-BC algorithm to its adjustable parameter $\alpha$. Then, the rate of convergence of the algorithm and its behavior on dealing with this major issue are discussed aligned with the other well-known meta-heuristic methods through several benchmark functions. \par

The adjustable parameters of the proposed approach are $num\_of\_stars$, determining how vast the exploration factor should be considered and   $\alpha$, the balancing probability between exploration and exploitation stages of the algorithm. On the one hand, more increase the value of $num\_of\_stars$ leads more exploration the space of solutions and achieving to probably more optimal solutions. On the other hand, more exploration of the solution space affects heavily on the time complexity of the algorithm. We have applied the value of the parameter $num\_of\_stars$ to $200$ based on empirical experiments and recommended settings in related works based on the trade of between the exploration of solution space and time complexity of the ME-BB-BC method. \par

 \begin{figure}[b!] 
 	\begin{subfigure}[b]{0.5\linewidth}
 		\centering
 		\includegraphics[width=1\linewidth]{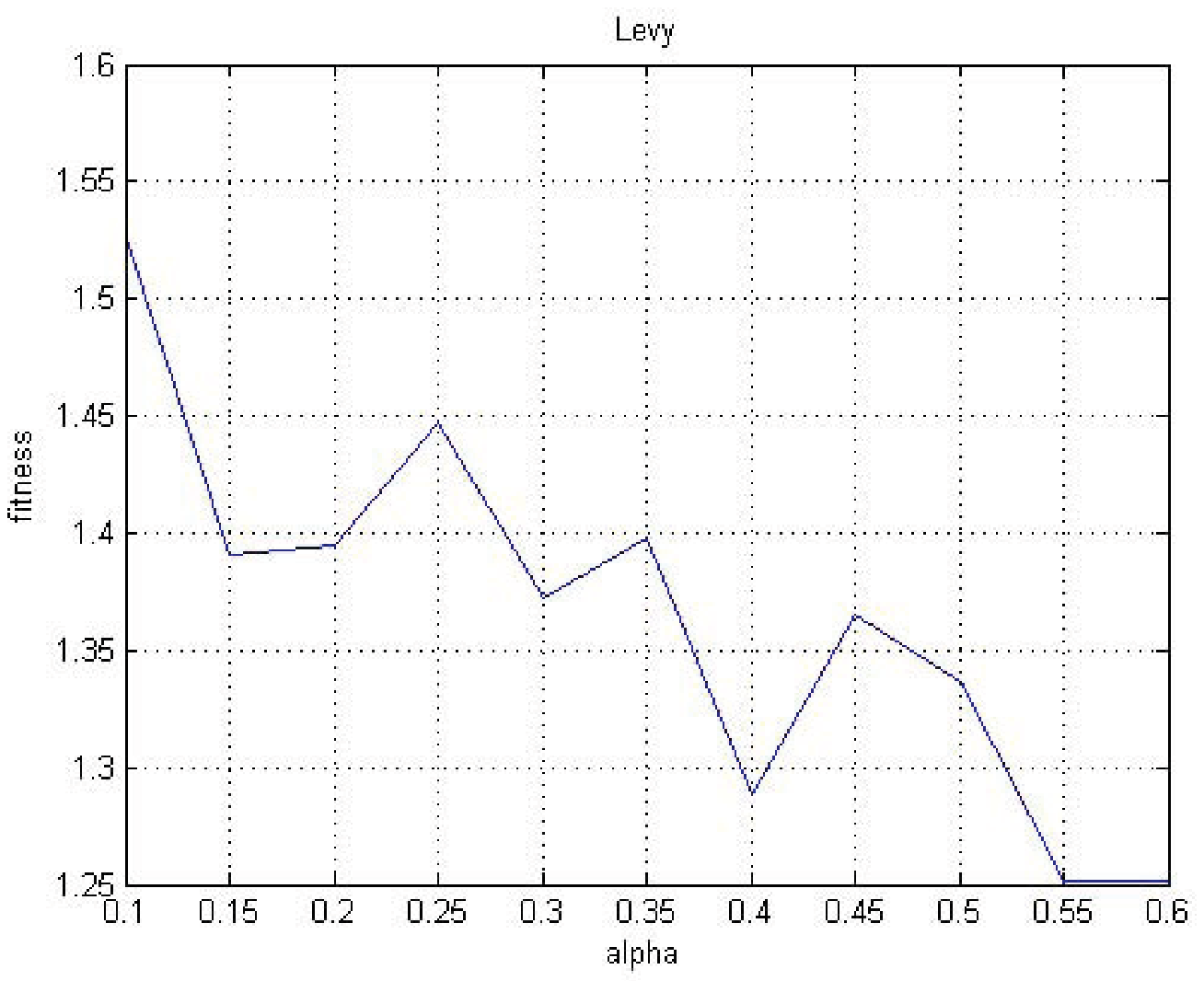} 
 		\caption{Levy} 
 		\label{fig:alphalevy} 
 		\vspace{4ex}
 	\end{subfigure}
 	\begin{subfigure}[b]{0.5\linewidth}
 		\centering
 		\includegraphics[width=1\linewidth]{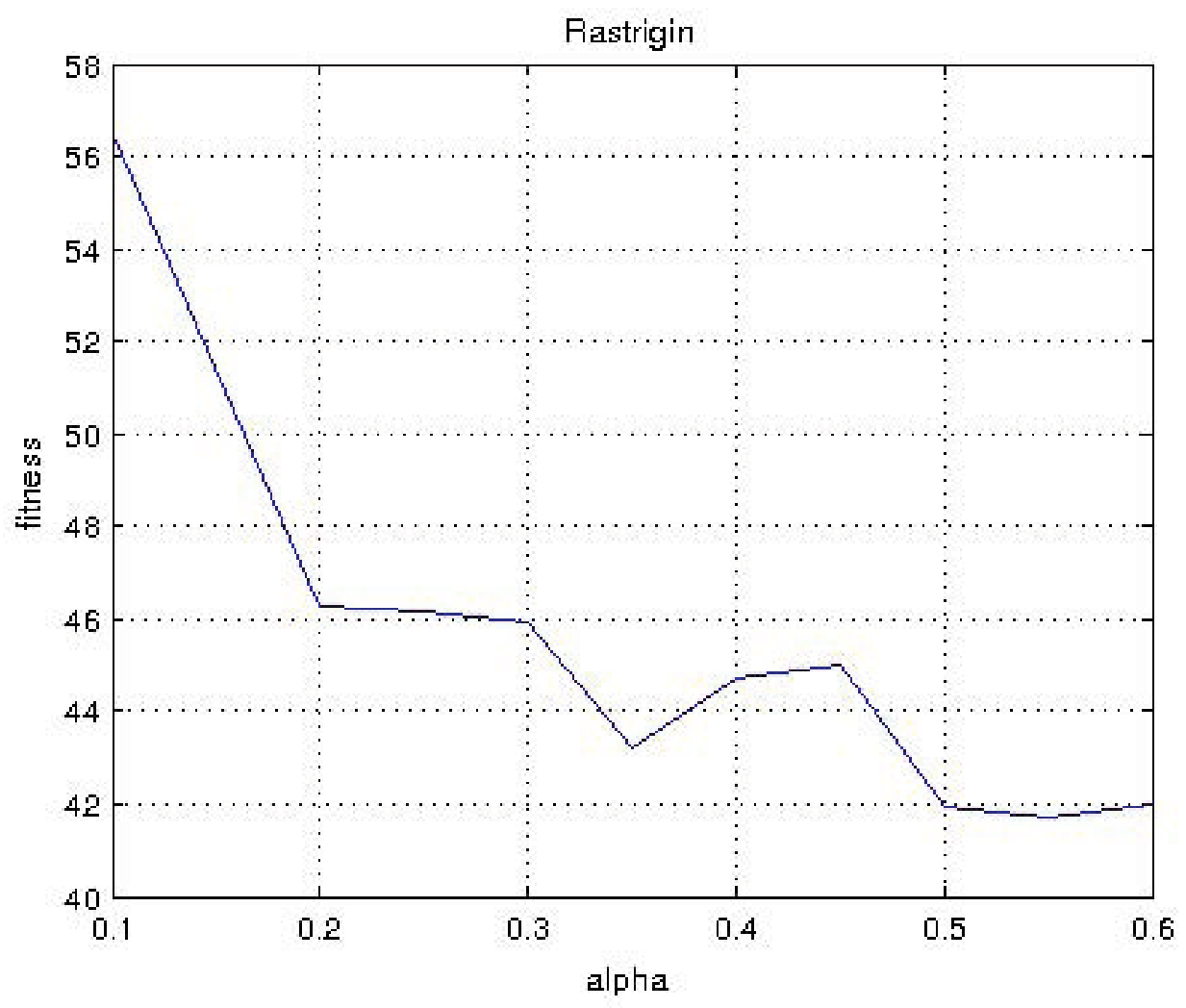} 
 		\caption{Rastrigin} 
 		\label{fig:alpharast} 
 		\vspace{4ex}
 	\end{subfigure} 
 	\caption{The sensitivity analysis of the ME-BB-BC algorithm to adjustable parameter $\alpha$ on (a) Levy, (b) Rastrigin functions.}
 	\label{fig:alphasens} 
 \end{figure}

The parameter $\alpha$ is initiated with small values in the beginning stages of the algorithm and is grown to larger amounts at the subsequent stages of the algorithm to balance between exploration and exploitation of the search strategy. Moreover, we have designed a careful sensitivity analysis over adjustable parameter $\alpha$ based on \emph{Levy} and \emph{Rastrigin} benchmark functions. The details of the results could be observed in Figure \ref{fig:alphasens}. Increasing the parameter  $\alpha$ generally results in better fitness values which is satisfied for both of  \emph{Levy} and \emph{Rastrigin} functions with the $50$ assumed dimension as observed in Sub-Figures \ref{fig:alphalevy} and \ref{fig:alpharast}.\par

\begin{figure}[h!] 
	\begin{subfigure}[b]{0.5\linewidth}
		\centering
		\includegraphics[width=1\linewidth]{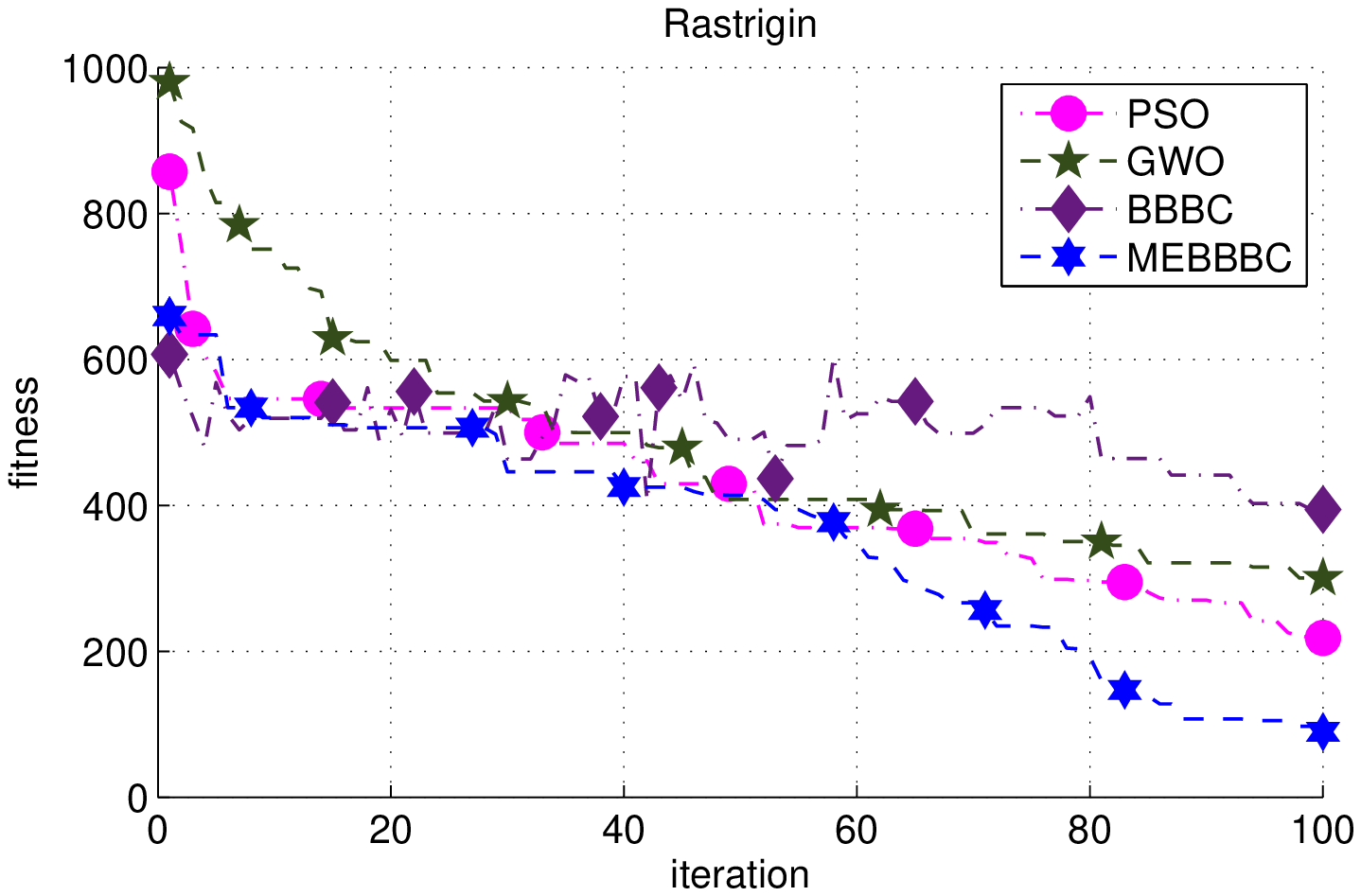} 
		\caption{Rastrigin} 
		\label{fig:convRast} 
		\vspace{4ex}
	\end{subfigure}
	\begin{subfigure}[b]{0.5\linewidth}
		\centering
		\includegraphics[width=1\linewidth]{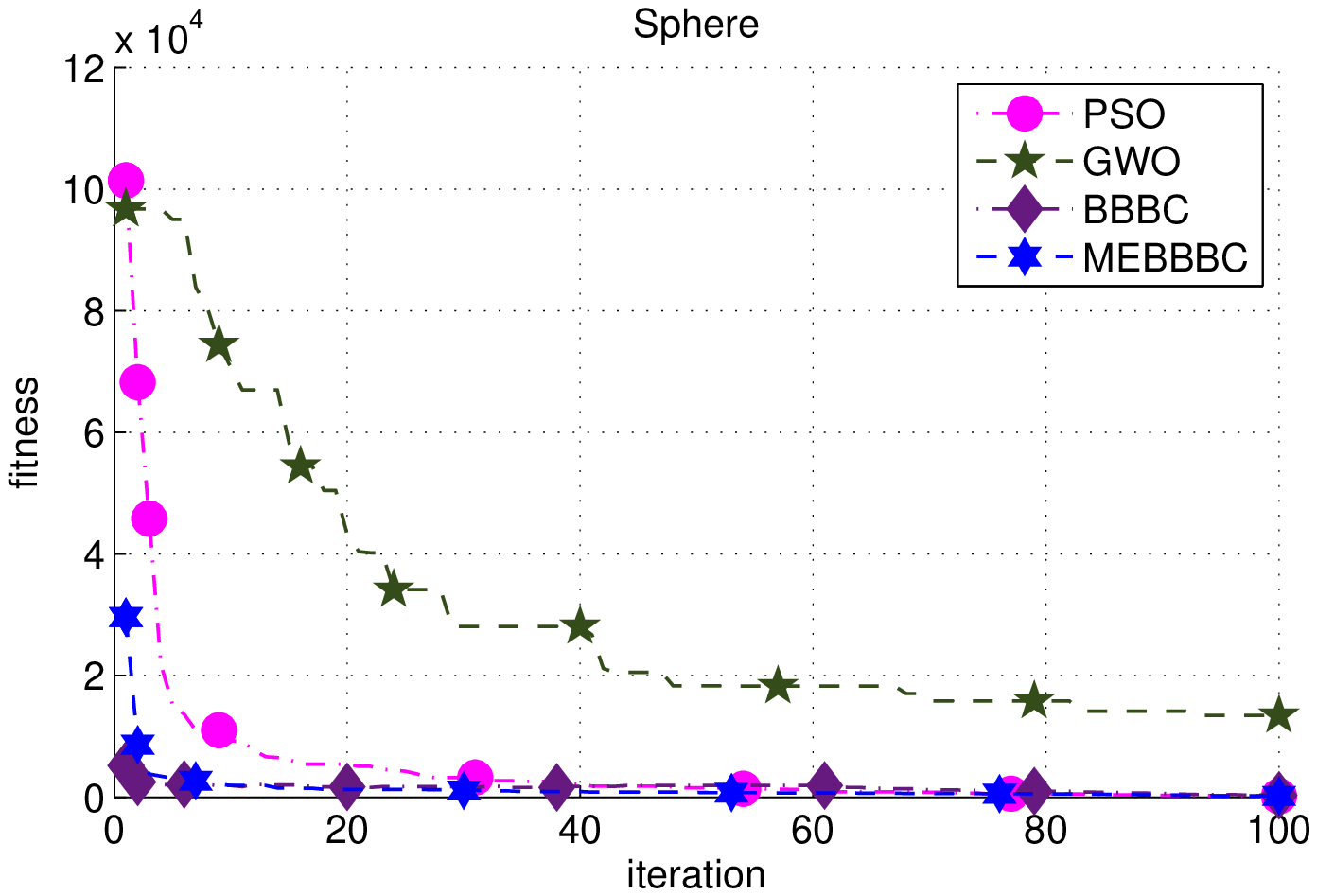} 
		\caption{Sphere} 
		\label{fig:convSpher} 
		\vspace{4ex}
	\end{subfigure} 
	\begin{subfigure}[b]{0.5\linewidth}
		\centering
		\includegraphics[width=1\linewidth]{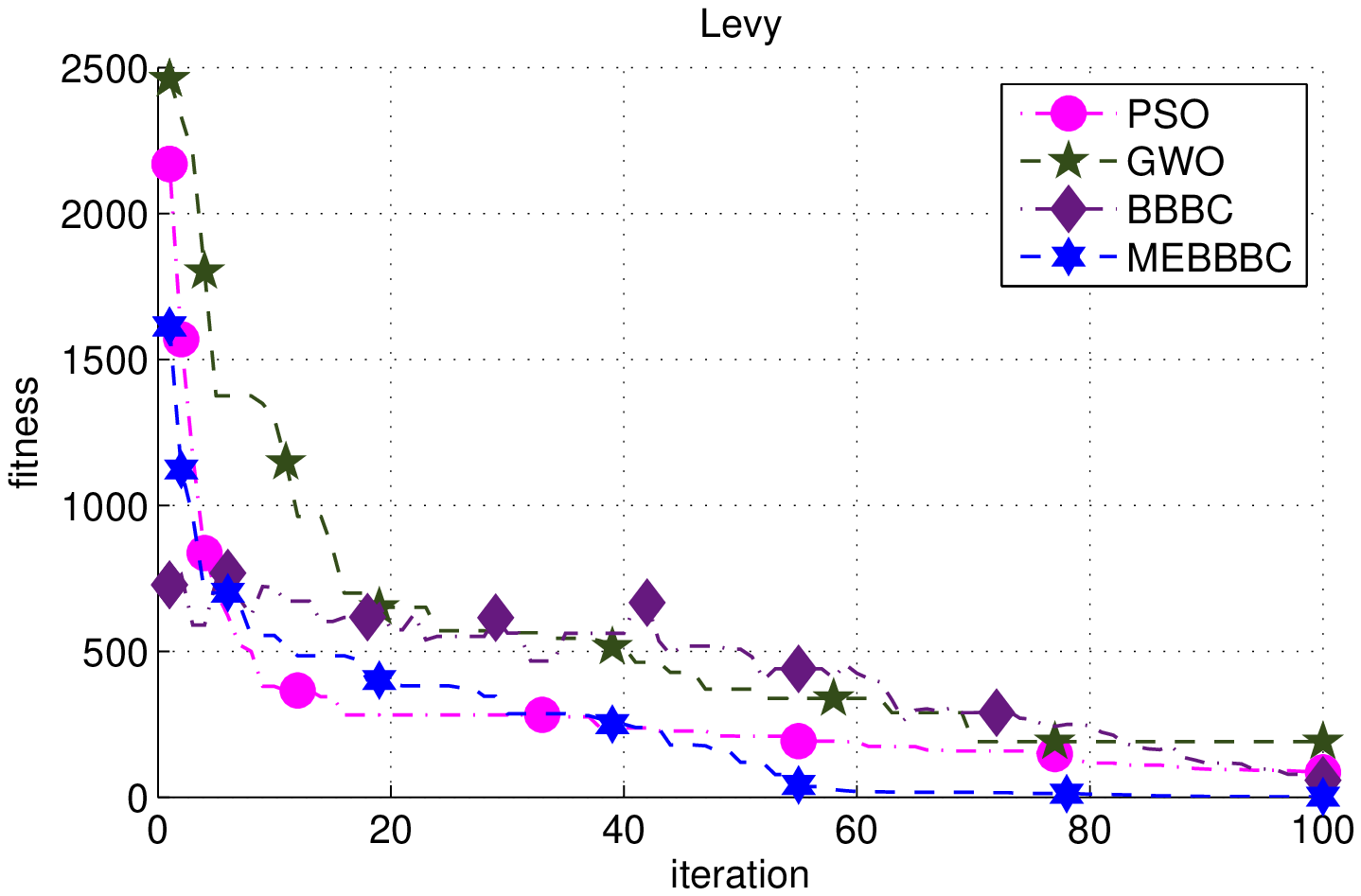} 
		\caption{Levy} 
		\label{fig:convLevy} 
	\end{subfigure}
	\begin{subfigure}[b]{0.5\linewidth}
		\centering
		\includegraphics[width=1\linewidth]{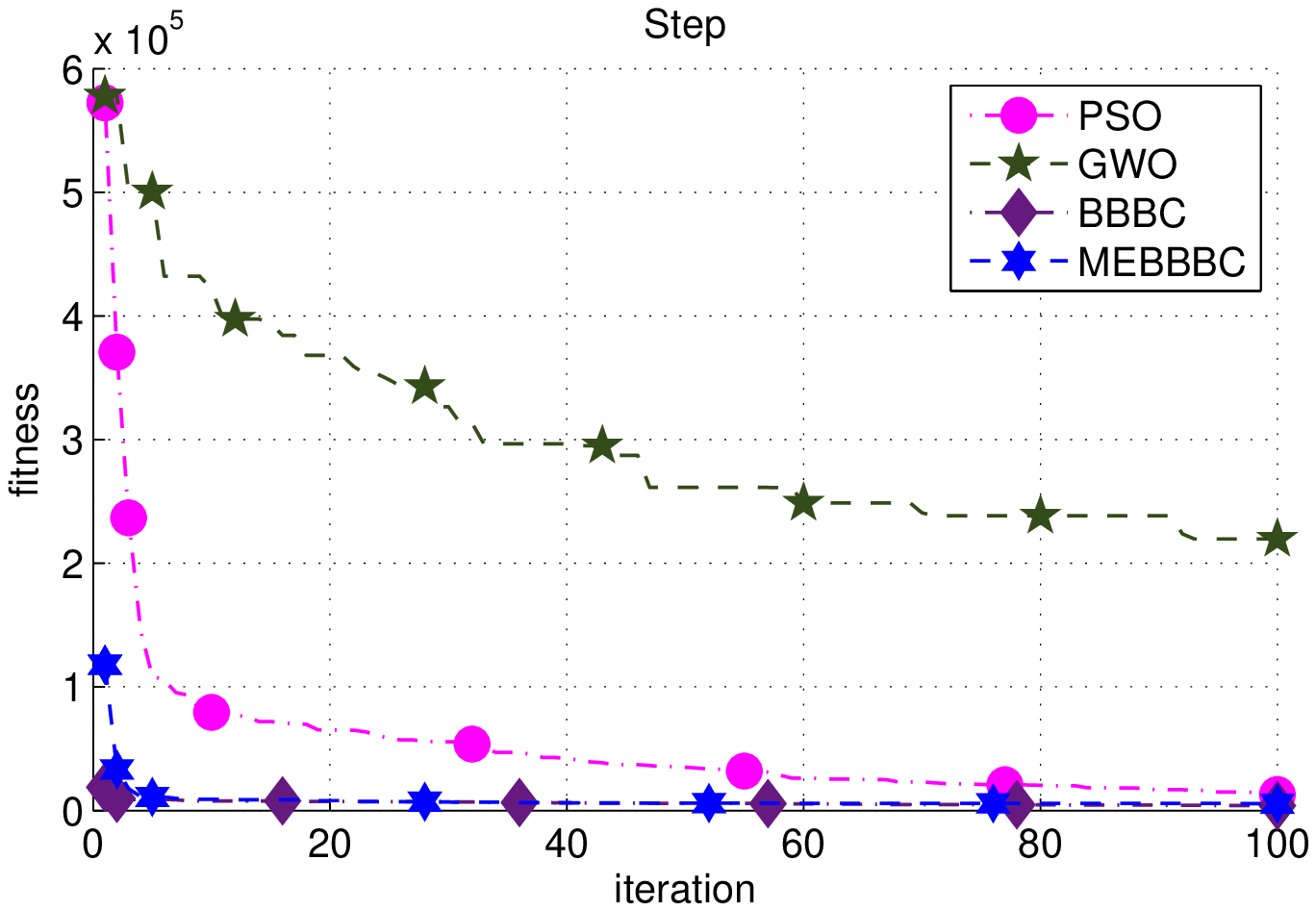} 
		\caption{Step} 
		\label{fig:convStep} 
	\end{subfigure} 
	\caption{Comparison of the results of the ME-BB-BC method with the \emph{PSO}, \emph{GWO}, and \emph{BBBC} algorithms on (a) Rastrigin, (b) Sphere, (c) Levy, and (d) Step benchmark functions.}
	\label{fig:convergence} 
\end{figure}

One of the general problems with the evolutionary algorithms is the slow rate of convergence. The ME-BB-BC algorithm contains a solution memory, and adjustable parameters  $\alpha$ and $num\_of\_stars$ to balance the rate of exploration and exploitation via using the solution memory where the rate of convergence of the algorithm depends on these two stages. We have compared the convergence of the proposed approach with the \emph{PSO}, Grey Wolf Optimizer \emph{GWO} \cite{mirjalili2014grey}, \emph{BB-BC} techniques  in Figure \ref{fig:convergence} based on the number of iterations versus the value of fitness function through four benchmark functions, \emph{Rastrigin}, \emph{Sphere}, \emph{Levy}, and \emph{Step}. The obtained results in Sub-Figures \ref{fig:convRast}, \ref{fig:convSpher}, \ref{fig:convLevy} and \ref{fig:convStep} indicate that the ME-BB-BC algorithm has better rate of convergence than the other algorithms and the decreasing pattern of the ME-BB-BC approach proves its less sensitivity to local optimums.
 
\section{The evaluation of the proposed approach}
\label{sec:evalproposedmeth}
To evaluate the ME-BB-BC, initially, a bunch of benchmark functions is used. Table~\ref{tbl:benchmarkfuncs} shows the benchmark functions where the optimal solution for all of them is zero. Furthermore, to more investigate the proposed algorithm and check the robustness of its results, statistical tests are performed on the obtained results.

\begin{table*}[t]
\caption{Specifications of Benchmark Functions}
\label{tbl:benchmarkfuncs}
\centering
\begin{tabular}{|c|c|c|c|}
\hline

Function &  Equation & Range & Solution\\
\hline\hline
Rastrigin & $\begin{aligned} \sum_{i=1}^n{(x_i^2 - 10cos(2\pi x_i) + 10)} \end{aligned}$ & [-5.12, 5.12] & [0, \dots, 0]\\

\hline\hline
Step & $\begin{aligned} \sum_{i=1}^n{([x_i+0.5])^2} \end{aligned}$ & [-100, 100] & [0, \dots, 0]\\
\hline\hline

Sphere & $\begin{aligned} \sum_{i=1}^n{x_i^2} \end{aligned}$ & [-100, 100] & [0, \dots, 0]\\
\hline\hline

Rosenbrock & $\begin{aligned} \sum_{i=1}^{n-1}{(100(x_{i+1} - x_i^2)^2 + (x_i-1)^2)} \end{aligned}$ & [-30, 30] & [0, \dots, 0]\\
\hline\hline

Zakharov & $\begin{aligned} \sum_{i=1}^nx_i^2 + (\sum_{i=1}^n0.5ix_i)^2 + (\sum_{i=1}^n0.5ix_i)^4 \end{aligned}$ & [-5, 10] & [0, \dots, 0]\\
\hline\hline

Levy & $ \begin{aligned} sin^2(\pi w_1)  + \sum_{i=1}^{n-1}(w_i-1)^2[1+10sin^2(\pi w_i +1)] \\+ (w_n-1)^2[1+sin^2(2\pi w_n)], \text{where} \\ 
w_i = 1+\frac{x_i-1}{4}, \text{for all } i=1, \dots, d \end{aligned} $ & [-15, 30] & [0, \dots, 0]\\
\hline\hline

Dixon Price & $\begin{aligned} (x_1-1)^2 + \sum_{i=2}^ni(2x_i^2 - x_{i-1})^2  \end{aligned}$ & [-10, 10] & [0, \dots, 0]\\

\hline

\end{tabular}
\end{table*}

\begin{table*}[h]
	\caption{(B)est, (A)verage, and (S)tandard Deviation of Algorithms on benchmark functions for 50 consecutive run of each algorithm.}
	\label{tbl:results}
	\centering
	\begin{tabular}{|c|c|ccccc|}
		\hline
		Function & V &  Proposed & GA & PSO & GWO & BB-BC\\
		\hline\hline
		
		\multirow{3}{*}{Rastrigin} & B & \textbf{39.18} & 51.33 & 78.11 & 40.94 & 269.79\\
		\hhline{~-}
		~ & A & \textbf{49.51} & 69.84 & 147.88 & 85.41 & 364.19 \\
		\hhline{~-}
		~ & S & \textbf{6.13} & 7.35 & 28.45 & 34.33 & 45.53 \\
		\hline\hline
		
		\multirow{3}{*}{Step} & B & \textbf{64.00} & 1308.00 & 94.00 & 65.00 & 98.00\\
		\hhline{~-}
		~ & A & \textbf{108.62} & 2420.50 & 244.64 & 155.20 & 323.14 \\
		\hhline{~-}
		~ & S & \textbf{20.47} & 403.50 & 79.47 & 73.14 & 163.69 \\
		\hline\hline
		
		\multirow{3}{*}{Sphere} & B & \textbf{47.10} & 1716.86 & 74.23 & 50.14 & 72.12\\
		\hhline{~-}
		~ & A & \textbf{69.77} & 2433.84 & 202.69 & 111.36 & 304.27 \\
		\hhline{~-}
		~ & S & \textbf{12.62} & 431.79 & 50.84 & 57.68 & 200.23 \\
		\hline\hline
		
		\multirow{3}{*}{Rosenbrock} & B & \textbf{245.41} & 472156.77 & 3021.37 & 247.13 & 251.41\\
		\hhline{~-}
		~ & A & \textbf{504.23} & 975412.12 & 10188.67 & 1152.74 & 1413.61 \\
		\hhline{~-}
		~ & S & \textbf{114.52} & 252115.63 & 4047.98 & 766.93 & 1099.13 \\
		\hline\hline
		
		\multirow{3}{*}{Zakharov} & B & \textbf{52.01} & 426.34 & 121.20 & 72.21 & 172.47\\
		\hhline{~-}
		~ & A & \textbf{100.65} & 586.10 & 243.32 & 179.05 & 349.08 \\
		\hhline{~-}
		~ & S & \textbf{33.82} & 54.32 & 80.58 & 59.31 & 99.23 \\
		\hline\hline
		
		\multirow{3}{*}{Levy} & B & \textbf{0.51} & 23.85 & 8.25 & 1.07 & 18.12\\
		\hhline{~-}
		~ & A & \textbf{0.95} & 33.68 & 50.82 & 4.50 & 195.45 \\
		\hhline{~-}
		~ & S & \textbf{0.21} & 6.85 & 33.58 & 2.80 & 31.74 \\
		\hline\hline
		
		\multirow{3}{*}{Dixon Price} & B & \textbf{5.15} & 3904.74 & 65.24 & 5.30 & 9.67\\
		\hhline{~-}
		~ & A & \textbf{12.34} & 9823.17 & 156.23 & 19.10 & 54.24 \\
		\hhline{~-}
		~ & S & \textbf{2.04} & 2922.37 & 59.50 & 9.04 & 34.20 \\
		\hline
		
	\end{tabular}
\end{table*}
The proposed algorithm is compared to Genetic Algorithm(GA), Particle Swarm Optimization(PSO), Grey Wolf Optimizer(GWO), and original BB-BC algorithm. Each algorithm has been run 50 times for each benchmark function and average, best and standard deviation of costs has been reported. In this study, the dimension of the benchmark functions is set to $50$ and the maximum number of iterations is $100$. Table \ref{tbl:results} shows the experimental results. The ME-BB-BC has performed the optimization task efficiently as compared to the other methods based on the cost function evaluation metrics. Moreover, lower standard deviation pointed us that the optimization algorithm converges to close results in different runs of the algorithm.

The nonparametric Friedman's statistical test is used to the 30 different runs of the proposed algorithm on Rosenbrock, Rastrigin, Sphere, and Step benchmark function with 50 dimension to investigate whether there exist significant differences among results or not. Table \ref{tbl:friedman} presents output from Friedman's test. Application of Friedman's test indicates that there is no statistically significant difference between the obtained results over the different runs of the proposed algorithm.

To further investigate the results of the proposed algorithm, t-test was applied in order to point out the difference between results of the ME-BB-BC and GWO, as best algorithm among other heuristic methods, Table \ref{tbl:ttest} presents the t-test compare of ME-BB-BC algorithm versus GWO algorithm over Rosenbrock, Rastrigin, Sphere, and Step benchmark functions. These results indicate that there is a significance difference between results of the ME-BB-BC and GWO algorithm.

\begin{table}[!ht]
	\centering
	\large
	\caption{Results of the non-parametric Friedman's statistical test}
	\label{tbl:friedman}
	\begin{tabular}{lc}
		\toprule[1.5pt]
		Benchmark Function & Significance  \\ 
		\midrule
		Rosenbrock & 0.38  \\ 
		
		Rastrigin & 0.49  \\
		
		Sphere & 0.71  \\
		
		Step & 0.33  \\
		\bottomrule[1.5pt]

	\end{tabular} 
\end{table}

\begin{table}[!h]
  \centering
  \large
  \caption{Results of the t-test between ME-BB-BC algorithm and GWO algorithm}
  \label{tbl:ttest}
  \begin{tabular}{lc}
    \toprule[1.5pt]
    Benchmark Function & p-value  \\ 
    \midrule
    Rosenbrock & 0.042  \\ 
    
    Rastrigin & 0.031  \\
    
    Sphere & 0.015  \\
    
    Step & 0.013  \\
    \bottomrule[1.5pt]

  \end{tabular} 
\end{table}

\section{Clustering using ME-BB-BC}
\label{sec:clustering_MBBBBC}

Due to the finding the interesting patterns from the dataset in the clustering situations, the meta-heuristic techniques are used extensively to uncover the hidden clusters from the dataset. Some advantages of these algorithms to perform clustering task can be mentioned such as less sensitivity to the initial starting points, reduction of  the amount of computation, and discover clusters with arbitrary shapes \cite{nanda_survey_2014}. \par

In this paper memory enriched big bang-big crunch algorithm proposed for clustering task. Besides, to improve the traditional clustering algorithms like k-means method, the proposed approach performs the clustering task in an efficient way. The proposed clustering algorithm operates in a four main steps that each will be illustrated in the following.
\begin{enumerate}
\item {\textbf{Starting Point:}}
\label{StartMEBB}
The proposed algorithm starts with an initial answer as a starting point of the procedure. This answer  consists of a vector of centers generated randomly in the range of allowable values. The example of a candidate solution for a clustering problem is given in Figure \ref{fig:centers}. 

\begin{figure}[!h]
\centering
\includegraphics[scale=.65]{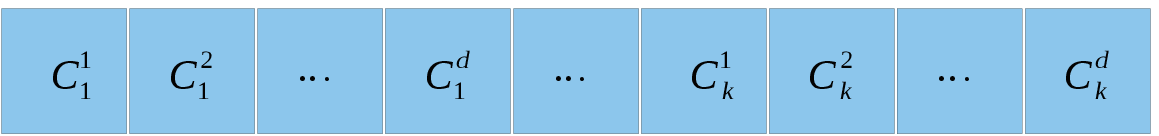}
\caption{The candidate centers for clustering where $k$ is the no. of clusters and  each sample of data has $d$ attributes.}
\label{fig:centers}
\end{figure}

\item {\textbf{Big Bang process:}}
\label{BigMEBB}
The initial solution is given to the Big Bang phase of the proposed algorithm to generate the particles in the solution space.

\item {\textbf{Evaluation:}}
\label{EvalMEBB}
The generated particles around the starting point should be tested based on equation \eqref{eqn:costfcn1} as follows,

\begin{equation}
\label{eqn:costfcn1}
Fitness = \sum_{n=1}^{N}{\sum_{k=1}^{K}I_{n,k}(d_n)\times(||d_n - \textrm{BC}_k||^2)} 
\end{equation}

where the $d_n$ is the n-th row of the dataset, $\textrm{BC}_k$ is the center of the k-th cluster derived from the Big Bang phase of the algorithm, and $I_{n, k}(d_n) $ equals to $1$ if $d_n$ belongs to k-th cluster and $0$ otherwise.

\item {\textbf{Big Crunch process:}}
\label{CrunchMEBB}
In this phase, the proposed algorithm compute  the center of the mass of randomly generated particles and use it as starting point of the algorithm in the next step. 
\end{enumerate}

If the stopping criteria of the algorithm are satisfied including reaching the maximum number of iterations, or fixed centers during consecutive iterations, the proposed algorithm will stop and the best solution among all solutions is reported. Otherwise, above steps will be repeated until the stopping conditions are satisfied. The details are presented in Algorithm \ref{alg:proposed_alg}.

\begin{algorithm}[!h]
\caption{The proposed clustering method based on memory enriched BB-BC Algorithm}
\label{alg:proposed_alg}
\begin{algorithmic}[1]
\renewcommand{\algorithmicrequire}{\textbf{Input:}}
\renewcommand{\algorithmicensure}{\textbf{Output:}}
\Require fitness function, memory size, number of stars, number of clusters
\Ensure  optimal cluster centers
 	
\\ \textit{Initialisation}:
\State $solution\_memory$ = memory with size $memory\_size$
\State $dim$ = dimension of clusters
\State $k$ = number of clusters
\State $centers = (c_1^1, \dots, c_1^{dim}, \dots,c_k^1, \dots, c_k^{dim})$ \Comment{randomly generate clusters centers with respect to variable range limitations}
\State  $\alpha= 0.1$ \Comment{memory selection rate}
\State $num\_of\_stars$ = number of stars

\Repeat
		\\ \textit{Big Bang Phase}:\Comment{create mass around starting point}
		\For{$i = 1$ to $num\_of\_stars$}
		\For{$j = 1$ to $dim$}
		\If{$rand(0,1) <= \alpha$ }
		\State $idx = rand([1,..., $memory\_size$])$ 
		\State $mass[i,j] = solution\_memory[idx, j]$ \Comment{select from memory}
		\Else
		\State $mass[i,j] = $ generate a star based on equ. \ref{eqn:bigbang}
		\EndIf
		\EndFor
		\EndFor
	
	\For{\textbf{each} star $\in mass$}
	\State $mass\_fitness[star] = fitness(star)$ based on equ. \ref{eqn:costfcn}
	\EndFor
	
	\\ \textit{Big Crunch Phase}:
		\State $c.o.m$ = calculate center of mass (equ. \ref{eqn:centerofmass})
	
		\If{$solution\_memory$ is not full}
		\State add $c.o.m$ into $solution\_memory$
		\Else 
		\State $worse$ = find the worst solution in $solution\_memory$
		\If{$fitness(c.o.m) > fitness(worse)$}
		\State remove $worse$ from the $solution\_memory$
		\State add $c.o.m$ into $solution\_memory$
		\EndIf
		\EndIf
		
		\State $centers = c.o.m$ \Comment{update}
		\State $\alpha = \alpha + 0.01\,\times\,\alpha$ \Comment{update}
		
	\Until max number of iterations or convergence
\end{algorithmic}
\label{alg:proposedclusteringmethod}
\end{algorithm}

\section{Experimental results}
\label{sec:exp_res}

The proposed algorithm on several standard datasets is examined where the description of the dataset are given in the Subsection \ref{data}. Then the obtained results are presented and discussed in Subsection \ref{results}.  

\subsection{Standard datasets}
\label{data}
The proposed method is experimentally evaluated using several standard datasets, including Iris, Wine, CMC, and Vowel. These datasets have been employed in many works and can be achieved from UCI Machine Learning Repository \cite{uciLichman2013}. The properties of the datasets are summarized in Table~\ref{tbl:data}. The detail for each of these datasets is as follows.

\begin{table}[h!]
  \centering
  \large
  \caption{summarized characteristic of real datasets}
  \label{tbl:data}
  \begin{tabular}{lccccc}
    \toprule[1.5pt]
    Dataset & \#Objects & \#Features & \#Clusters \\ 
    \midrule 
    Iris & 150 & 4 & 3(50, 50, 50) \\ 
    Wine & 178 & 13 & 3(59, 71, 48) \\
    CMC  & 1473 & 9 & 3(629, 334, 510)\\
    Vowel & 871 & 3 & 6(72, 89, 172, 151, 207, 180)\\
    Glass & 214 & 9 & 6(70, 76, 17, 13, 9, 29)\\
    Cancer & 638 & 9 & 2(444, 239) \\
    \bottomrule[1.5pt]
    
  \end{tabular} 
\end{table}

\begin{itemize}
\item \textbf{Iris:} This dataset contains 150 instances of iris plants, with 4 attributes and is divided to 3 categories each cluster contains 50 objects.
	
\item \textbf{Wine:} This data is the consequences of investigation of wines developed in the same area in Italy yet got from three unique cultivars. The analysis determines the amounts of 13 constituents found in each of the 3 sorts of wines. This dataset contains investigations of 178 instances.
	
\item \textbf{CMC:} This dataset is a subset of the 1987 National Indonesia Contraceptive Prevalence Survey. The instances are married women who were either not pregnant or don't know whether they were or not. The issue is to foresee the present preventative technique decision of a woman considering her demographic and financial attributes. This dataset contains 1473 items with 9 attributes and 3 clusters.
	
\item \textbf{Vowel:} The Vowel dataset comprises of 871 Indian Telugu vowel sounds. There are 6 overlapping classes and 3 features. Vowel dataset contains samples with low, medium and high dimensions.
	
\item \textbf{Glass:} Glass dataset has 214 instances with 9 features. The dataset has 6 unique clusters of different sort of windows.
	
\item \textbf{Cancer:} Wisconsin Breast Cancer Dataset has 683 samples with 9 traits. malignant and benign are two clusters which instances of this dataset falls into.
	
\end{itemize}

\subsection{Results}
\label{results}
The proposed algorithm is compared to the Genetic Algorithm (GA), Particle Swarm Optimization (PSO), Grey Wolf Optimizer (GWO), original Big Bang-Big Crunch algorithm (BB-BC) and k-means algorithm. It should be noted that all algorithms have been implemented in MATLAB and their parameters are being set according to their reference paper. Each algorithm has been run for 50 consecutive times on system with Windows 7, 4 Gigabyte of RAM and core i5 2.66 GHz processor. Table~\ref{tbl:cluster} presents best, average and standard deviation of different runs of applying these algorithms for given datasets. According to equation \ref{eqn:costfcn}, the lowest the value of fitness function is, the better the clustering quality will be. By all it means the lower value for sum of cluster distance is more desirable. Lower standard deviation indicates that the optimization algorithm converges to close results in different runs of the algorithm. The experimental results have shown us better performance of the ME-BB-BC algorithm as compared to the other methods.

\begin{table}[!h]
\caption{(B)est, (A)verage, and (S)tandard Deviation of Algorithms on real world clustering datasets for 50 consecutive run of each algorith.}
\label{tbl:cluster}
\centering
\begin{tabular}{|c|c|cccccc|}
\hline
Dataset & V &  Proposed & GA & PSO & GWO & BB-BC & KMeans\\
\hline\hline

\multirow{3}{*}{Iris}      & B & \textbf{96.60} & 96.72 & 96.65 & 97.10 & 96.91 & 97.32\\
\hhline{~-}
					  ~ & A & \textbf{96.75} & 97.30 & 96.81 & 120.12 & 97.23 & 98.86\\
\hhline{~-}
		   ~ & S & \textbf{0.21} & 0.47 & 0.21 & 11.28 & 0.82 & 6.11\\
\hline\hline

\multirow{3}{*}{Wine}      & B & \textbf{16292.20} & 16294.16 & 16292.79 & 16292.52 & 16363.96 & 16555.67 \\
\hhline{~-}
					  ~ & A & \textbf{16293.12} & 16301.55 & 16293.74 & 16297.23 & 17013.95 & 17684.44 \\
\hhline{~-}
		   ~ & S & \textbf{0.69} & 8.53 & 1.02 & 3.86 & 880.99 & 930.98 \\
\hline\hline

\multirow{3}{*}{CMC}      & B & \textbf{5532.25} & 5536.20 & 5532.49 & 5548.27 & 5852.47 & 5542.18\\
\hhline{~-}
					  ~ & A &\textbf{ 5532.49} & 5543.04 & 5533.96 & 5571.98 & 6213.94 & 5543.69 \\
\hhline{~-}
		   ~ & S & \textbf{0.26} & 3.97 & 3.93 & 14.91 & 584.78 & 1.57\\
\hline\hline

\multirow{3}{*}{Vowel}      & B & 148983.31 & 149382.97 & \textbf{148970.79} & 150030.76 & 169551.16 & 149383.99 \\
\hhline{~-}
					  ~ & A & \textbf{149440.43} & 151001.32 & 149461.66 & 150430.72 & 194489.91 & 154259.91 \\
\hhline{~-}
		   ~ & S & \textbf{452.01} & 1479.30 & 501.79 & 445.37 & 24402.14 & 4215.43 \\
\hline\hline

\multirow{3}{*}{Glass}      & B & \textbf{213.16} & 223.512 & 222.55 & 218.35 & 440.31 & 214.24\\
\hhline{~-}
					  ~ & A & \textbf{227.00} & 239.33 & 247.36 & 241.85 & 664.18 & 229.34\\
\hhline{~-}
		   ~ & S & \textbf{10.09} & 7.45 & 10.45 & 11.22 & 68.93 & 14.29\\
\hline\hline

\multirow{3}{*}{Cancer}      & B & \textbf{2964.39} & 2966.44 & 2967.96 & 2968.28 & 4142.24 & 2986.96 \\
\hhline{~-}
					  ~ & A & \textbf{2964.45} & 2970.55 & 2980.00 & 3000.14 & 5472.48 & 2988.22 \\
\hhline{~-}
		   ~ & S & \textbf{0.03} & 3.00 & 18.92 & 26.57 & 624.84 & 0.51 \\
\hline

\end{tabular}
\end{table}

Furthermore, to statistically investigate the obtained results and analyze the robustness of results of the proposed algorithm over different runs of the algorithm, non-parametric Friedman's test is applied to the 30 different runs of the proposed clustering approach on clustering datasets. Table \ref{tbl:clsfriedman} presents output of Friedman's test. Application of Friedman's test indicates that there is no statistically significant difference between the obtained results over the different runs of the proposed clustering algorithm.

\begin{table}[!h]
  \centering
  \large
  \caption{Results of the non-parametric Friedman's statistical test}
  \label{tbl:friedman-cls}
  \begin{tabular}{lc}
    \toprule[1.5pt]
    Dataset & Significance  \\ 
    \midrule
    Iris & 0.98  \\
    
    Glass & 0.99  \\ 
    
    CMC & 0.99  \\
    
    Vowel & 0.99  \\
    
    Glass & 0.69  \\
    
    Cancer & 0.99  \\
    \bottomrule[1.5pt]

  \end{tabular} 
\end{table}

\begin{table}[!h]
	\caption{(B)est, (A)verage, and (S)tandard Deviation of Algorithms on real world clustering datasets for 50 consecutive run of each algorithm.}
	\label{tbl:me-kmeans}
	\centering
	\begin{tabular}{|c|c|cc|}
		\hline
		Dataset & V &  ME-BB-BC & kMEBB\\
		\hline\hline
		
		\multirow{3}{*}{Iris}      & B & \textbf{96.60} & 96.66\\
		\hhline{~-}
		~ & A & \textbf{96.75} & 97.41\\
		\hhline{~-}
		~ & S & \textbf{0.21} & 0.51\\
		\hline\hline
		
		\multirow{3}{*}{Wine}      & B & \textbf{16292.20} & 16292.21 \\
		\hhline{~-}
		~ & A & 16293.12 & \textbf{16262.30}\\
		\hhline{~-}
		~ & S & \textbf{0.69} & 0.22\\
		\hline\hline
		
		\multirow{3}{*}{CMC}      & B & 5532.25 & \textbf{5532.15}\\
		\hhline{~-}
		~ & A & 5532.49 & \textbf{5532.25}\\
		\hhline{~-}
		~ & S & 0.26 & \textbf{0.15} \\
		\hline\hline
		
		\multirow{3}{*}{Vowel}      & B & \textbf{148983.31} & 148990.80 \\
		\hhline{~-}
		~ & A & \textbf{149440.43} & 149445.96\\
		\hhline{~-}
		~ & S & 452.01 & \textbf{450.21}\\
		\hline\hline
		
		\multirow{3}{*}{Glass}      & B & \textbf{213.16} & 214.21\\
		\hhline{~-}
		~ & A & 227.00 & \textbf{226.65} \\
		\hhline{~-}
		~ & S & \textbf{10.09} & 10.43 \\
		\hline\hline
		
		\multirow{3}{*}{Cancer}      & B & \textbf{2964.39} & 2964.87  \\
		\hhline{~-}
		~ & A & \textbf{2964.45} & 2965.24 \\
		\hhline{~-}
		~ & S & \textbf{0.03} & 0.10  \\
		\hline
		
	\end{tabular}
\end{table}

Also the k-means algorithm can be combined with  meta-heuristic algorithms for clustering applications. In this regard, we have designed a hybrid clustering algorithm based on a slight change in the ME-BB-BC algorithm aligned with k-means named as \emph{kMEBB} where the step 3 of the original algorithm in Section \ref{sec:clustering_MBBBBC} is modified by applying the typical k-means procedure before the final evaluation stage for further improvement of the generated solutions of the Big Bang phase. 

We compare  the performance of the ME-BB-BC algorithm with the  hybrid kMEBB algorithm on the datasets  in Table \ref{tbl:data}. Based on the obtained results in Table \ref{tbl:me-kmeans}, ME-BB-BC performs equally or better than kMEBB on these datasets. These results reassured us that the  ME-BB-BC can be regarded as a proper choice for clustering applications among other meta-heuristic techniques.

\section{Conclusion and future work}
\label{sec:conc}

In this paper, via a memory enriched approach aligned to the classical BB-BC algorithm,  we are succeeded to enhance significantly its efficiency versus other standard meta-heuristic approaches for benchmark optimization functions and also clustering applications. Not only the deficiencies of the BB-BC method were alleviated through the smart way of using the memory of previously created solutions, but also these solutions were combined with newly candidate ones in a probabilistic random walk manner to improve the exploitation and exploration of the  proposed method. Furthermore,  this algorithm has been applied  for clustering aims. To evaluate the performance of the proposed algorithm, the experimental results were compared with other similar data clustering algorithms. Implementation results on benchmark functions and clustering datasets showed us the superiority of the proposed algorithm over other algorithms. There exist different directions to extend the proposed evolutionary algorithm for future works as the following.

\begin{itemize}

\item A hybrid clustering algorithm of big bang-big crunch and k-means which improves shortcomings of the k-means method.

\item The investigation of the proposed method aligned with statistical model based clustering algorithms. The learning parameters of the model based clustering paradigm are usually approximated through an iterative procedure named as Expectation Maximization (\emph{EM}) algorithm \cite{murphy_machine_2012}. While EM suffers from slow rate convergence, the ME-BB-BC approach would  alleviate this major problem as a future work on this field.

\item The proposed algorithm can also be used for multi objective optimization.

\item Application of ME-BB-BC in technical settings such as power dispatch systems.

\end{itemize}
\section*{Acknowledgments}
The authors would like to thank the anonymous reviewers for providing helpful comments and recommendations which improves the paper significantly.

\bibliographystyle{ieeetr}

\begin{thebibliography}{10}
	
	\bibitem{murphy_machine_2012}
	K.~Murphy, {\em Machine {Learning}: {A} {Probabilistic} {Perspective}}.
	\newblock Cambridge, Mass.: The MIT Press, 1 edition~ed., Aug. 2012.
	
	\bibitem{jain2010data}
	A.~K. Jain, ``Data clustering: 50 years beyond k-means,'' {\em Pattern
		recognition letters}, vol.~31, no.~8, pp.~651--666, 2010.
	
	\bibitem{gan2007data}
	G.~Gan, C.~Ma, and J.~Wu, {\em Data clustering: theory, algorithms, and
		applications}, vol.~20.
	\newblock Siam, 2007.
	
	\bibitem{ahmad2010cluster}
	N.~Ahmad, D.~Alahakoon, and R.~Chau, ``Cluster identification and separation in
	the growing self-organizing map: application in protein sequence
	classification,'' {\em Neural Computing and Applications}, vol.~19, no.~4,
	pp.~531--542, 2010.
	
	\bibitem{moftah2014adaptive}
	H.~M. Moftah, A.~T. Azar, E.~T. Al-Shammari, N.~I. Ghali, A.~E. Hassanien, and
	M.~Shoman, ``Adaptive k-means clustering algorithm for mr breast image
	segmentation,'' {\em Neural Computing and Applications}, vol.~24, no.~7-8,
	pp.~1917--1928, 2014.
	
	\bibitem{moradi_hybrid_2016}
	P.~Moradi and M.~Gholampour, ``A hybrid particle swarm optimization for feature
	subset selection by integrating a novel local search strategy,'' {\em Applied
		Soft Computing}, vol.~43, pp.~117--130, June 2016.
	
	\bibitem{lam2013pso}
	Y.-K. Lam, P.~W.-M. Tsang, and C.-S. Leung, ``Pso-based k-means clustering with
	enhanced cluster matching for gene expression data,'' {\em Neural Computing
		and Applications}, vol.~22, no.~7-8, pp.~1349--1355, 2013.
	
	\bibitem{dinu2014clustering}
	L.~P. Dinu and R.~T. Ionescu, ``Clustering based on median and closest string
	via rank distance with applications on dna,'' {\em Neural Computing and
		Applications}, vol.~24, no.~1, pp.~77--84, 2014.
	
	\bibitem{welch1982algorithmic}
	W.~J. Welch, ``Algorithmic complexity: three np-hard problems in computational
	statistics,'' {\em Journal of Statistical Computation and Simulation},
	vol.~15, no.~1, pp.~17--25, 1982.
	
	\bibitem{al1995tabu}
	K.~S. Al-Sultan, ``A tabu search approach to the clustering problem,'' {\em
		Pattern Recognition}, vol.~28, no.~9, pp.~1443--1451, 1995.
	
	\bibitem{krishna1999genetic}
	K.~Krishna and M.~N. Murty, ``Genetic k-means algorithm,'' {\em Systems, Man,
		and Cybernetics, Part B: Cybernetics, IEEE Transactions on}, vol.~29, no.~3,
	pp.~433--439, 1999.
	
	\bibitem{shelokar2004ant}
	P.~Shelokar, V.~K. Jayaraman, and B.~D. Kulkarni, ``An ant colony approach for
	clustering,'' {\em Analytica Chimica Acta}, vol.~509, no.~2, pp.~187--195,
	2004.
	
	\bibitem{cura2012particle}
	T.~Cura, ``A particle swarm optimization approach to clustering,'' {\em Expert
		Systems with Applications}, vol.~39, no.~1, pp.~1582--1588, 2012.
	
	\bibitem{zhang2010artificial}
	C.~Zhang, D.~Ouyang, and J.~Ning, ``An artificial bee colony approach for
	clustering,'' {\em Expert Systems with Applications}, vol.~37, no.~7,
	pp.~4761--4767, 2010.
	
	\bibitem{kao2008hybridized}
	Y.-T. Kao, E.~Zahara, and I.-W. Kao, ``A hybridized approach to data
	clustering,'' {\em Expert Systems with Applications}, vol.~34, no.~3,
	pp.~1754--1762, 2008.
	
	\bibitem{jordehi2015enhanced}
	A.~R. Jordehi, ``Enhanced leader pso (elpso): A new pso variant for solving
	global optimisation problems,'' {\em Applied Soft Computing}, vol.~26,
	pp.~401--417, 2015.
	
	\bibitem{jordehi2015enhancedPSO}
	A.~R. Jordehi, J.~Jasni, N.~A. Wahab, M.~Kadir, and M.~Javadi, ``Enhanced
	leader pso (elpso): a new algorithm for allocating distributed tcsc’s in
	power systems,'' {\em International Journal of Electrical Power \& Energy
		Systems}, vol.~64, pp.~771--784, 2015.
	
	\bibitem{zhang2015new}
	L.~Zhang, Y.~Tang, C.~Hua, and X.~Guan, ``A new particle swarm optimization
	algorithm with adaptive inertia weight based on bayesian techniques,'' {\em
		Applied Soft Computing}, vol.~28, pp.~138--149, 2015.
	
	\bibitem{tabakhi2014unsupervised}
	S.~Tabakhi, P.~Moradi, and F.~Akhlaghian, ``An unsupervised feature selection
	algorithm based on ant colony optimization,'' {\em Engineering Applications
		of Artificial Intelligence}, vol.~32, pp.~112--123, 2014.
	
	\bibitem{jordehi2015brainstorm}
	A.~R. Jordehi, ``Brainstorm optimisation algorithm (bsoa): An efficient
	algorithm for finding optimal location and setting of facts devices in
	electric power systems,'' {\em International Journal of Electrical Power \&
		Energy Systems}, vol.~69, pp.~48--57, 2015.
	
	\bibitem{jordehi2015seeker}
	A.~R. Jordehi, ``Seeker optimisation (human group optimisation) algorithm with
	chaos,'' {\em Journal of Experimental \& Theoretical Artificial
		Intelligence}, vol.~27, no.~6, pp.~753--762, 2015.
	
	\bibitem{heidari2015efficient}
	A.~A. Heidari, R.~A. Abbaspour, and A.~R. Jordehi, ``An efficient chaotic water
	cycle algorithm for optimization tasks,'' {\em Neural Computing and
		Applications}, pp.~1--29, 2015.
	
	\bibitem{jordehi2015chaotic}
	A.~R. Jordehi, ``A chaotic artificial immune system optimisation algorithm for
	solving global continuous optimisation problems,'' {\em Neural Computing and
		Applications}, vol.~26, no.~4, pp.~827--833, 2015.
	
	\bibitem{erol2006new}
	O.~K. Erol and I.~Eksin, ``A new optimization method: big bang--big crunch,''
	{\em Advances in Engineering Software}, vol.~37, no.~2, pp.~106--111, 2006.
	
	\bibitem{verma2013tlbo}
	H.~Verma and P.~Mafidar, ``Tlbo based voltage stable environment friendly
	economic dispatch considering real and reactive power constraints,'' {\em
		Journal of The Institution of Engineers (India): Series B}, vol.~94, no.~3,
	pp.~193--206, 2013.
	
	\bibitem{kucuktezcan2015preventive}
	C.~F. Kucuktezcan and V.~I. Genc, ``Preventive and corrective control
	applications in power systems via big bang--big crunch optimization,'' {\em
		International Journal of Electrical Power \& Energy Systems}, vol.~67,
	pp.~114--124, 2015.
	
	\bibitem{tang2010big}
	H.~Tang, J.~Zhou, S.~Xue, and L.~Xie, ``Big bang-big crunch optimization for
	parameter estimation in structural systems,'' {\em Mechanical Systems and
		Signal Processing}, vol.~24, no.~8, pp.~2888--2897, 2010.
	
	\bibitem{jordehi2014chaotic}
	A.~R. Jordehi, ``A chaotic-based big bang--big crunch algorithm for solving
	global optimisation problems,'' {\em Neural Computing and Applications},
	vol.~25, no.~6, pp.~1329--1335, 2014.
	
	\bibitem{xu2008clustering}
	R.~Xu and D.~Wunsch, {\em Clustering}, vol.~10.
	\newblock Wiley-IEEE Press, 2008.
	
	\bibitem{bishop_pattern_2007}
	C.~M. Bishop, {\em Pattern {Recognition} and {Machine} {Learning}}.
	\newblock Springer, 1st ed. 2006. corr. 2nd printing~ed., Oct. 2007.
	
	\bibitem{innovative_2013}
	B.~Xing and W.-J. Gao, {\em Innovative {Computational} {Intelligence}: {A}
		{Rough} {Guide} to 134 {Clever} {Algorithms}}.
	\newblock New York, NY: Springer, 2014 edition~ed., Dec. 2013.
	
	\bibitem{mahdavi_improved_2007}
	M.~Mahdavi, M.~Fesanghary, and E.~Damangir, ``An improved harmony search
	algorithm for solving optimization problems,'' {\em Applied Mathematics and
		Computation}, vol.~188, pp.~1567--1579, May 2007.
	
	\bibitem{shi_empirical_1999}
	Y.~Shi and R.~C. Eberhart, ``Empirical study of particle swarm optimization,''
	in {\em Proceedings of the 1999 {Congress} on {Evolutionary} {Computation},
		1999. {CEC} 99}, vol.~3, p.~1950 Vol. 3, 1999.
	
	\bibitem{mirjalili2014grey}
	S.~Mirjalili, S.~M. Mirjalili, and A.~Lewis, ``Grey wolf optimizer,'' {\em
		Advances in Engineering Software}, vol.~69, pp.~46--61, 2014.
	
	\bibitem{nanda_survey_2014}
	S.~J. Nanda and G.~Panda, ``A survey on nature inspired metaheuristic
	algorithms for partitional clustering,'' {\em Swarm and Evolutionary
		Computation}, vol.~16, pp.~1--18, June 2014.
	
	\bibitem{uciLichman2013}
	M.~Lichman, ``{UCI} machine learning repository.''
	http://archive.ics.uci.edu/ml, 2013.
	
\end{thebibliography}

\end{document}